%% file: main.tex
\documentclass{article}
\usepackage[linesnumbered,ruled]{algorithm2e}
\usepackage{spconf,amsmath,graphicx}
\usepackage{url}
\usepackage{makecell}
\usepackage{booktabs}  
\usepackage{float}
\usepackage{multirow}
\usepackage{amssymb}
\ninept
\DeclareMathOperator*{\argmax}{arg\,max}

\title{High-Precision Voice Search Query Correction via Retrievable~Speech-Text Embeddings}
\makeatother
\name{\begin{tabular}{c} Christopher Li, Gary Wang, Kyle Kastner, Heng Su, Allen Chen,Andrew Rosenberg \\
Zhehuai Chen\sthanks{Work done while at Google}, Zelin Wu, Leonid Velikovich, Pat Rondon, Diamantino Caseiro, Petar Aleksic\end{tabular}}
\address{Google, Inc.}
\begin{document}

\maketitle
\begin{abstract}
Automatic speech recognition (ASR) systems can suffer from poor recall for various reasons, such as noisy audio, lack of sufficient training data, etc.
Previous work has shown that recall can be improved by
retrieving rewrite candidates from a large database of likely, contextually-relevant alternatives to the hypothesis text using nearest-neighbors search over embeddings of the ASR hypothesis text to correct and candidate corrections.
However, ASR-hypothesis-based retrieval can yield poor precision if the textual hypotheses are too phonetically dissimilar to the transcript truth.
In this paper, we eliminate the hypothesis-audio mismatch problem by querying the correction database directly using embeddings derived from the utterance audio; the embeddings of the utterance audio and candidate corrections are produced by multimodal speech-text embedding networks trained to place the embedding of the audio of an utterance and the embedding of its corresponding textual transcript close together.
After locating an appropriate correction candidate using nearest-neighbor search, we score the candidate  with its speech-text embedding distance before adding the candidate to the original n-best list.
We show a relative word error rate (WER) reduction of 6\% on utterances whose transcripts appear in the candidate set, without increasing WER on general utterances.

\end{abstract}
\input{introduction}
\input{related_work}

\input{system_design}
\input{inference}
\input{experiments}
\input{results}
\input{conclusion}
\input{acknowledgements}
\bibliographystyle{IEEEbib}
\bibliography{main}
\end{document}

%% file: introduction.tex
\section{Introduction}
\label{sec:intro}
ASR contextualization (or biasing) systems improve recognition accuracy for queries when
additional information is available, e.g., the dialog state, device state,
or the language of the query.  For example, the correct hypothesis
for a query to a voice assistant device in a music-playing state is more likely to
contain musician ``Eminem'' than the
acoustically identical, but linguistically incorrect, alternative ``M\&M.'' 

ASR correction is a modular approach to ASR contextualization which allows for the use of the full ASR output and other resources which may not be available, or may be too expensive to access, during first-pass recognition.
% TODO(rondon): Move more of the background about retrieval here
% Talk about dual encoders and fast nearest-neighbors search enabling correction using large databases
% Tradeoff: working from the text, plus a large database, makes it more likely that (1) the embedding will not accurately reflect the utterance audio and (2) there will be a false positive match in the database.
In particular, in many domains, it is desirable to use very large collections of correction candidates --- for example, for media queries, we can take advantage of large databases of artists, songs, and albums.
We may quickly find correction candidates in a large database as follows:
\begin{enumerate}
    \item A ``dual encoder'' model consisting of two encoder networks (or a single shared encoder) is trained to map the target for correction (e.g., ASR hypothesis text) and candidate corrections (e.g., database entries) into the same embedding space (as in, e.g.,~\cite{flare_entity_retrieval}), such that incorrect text is close to the corresponding correction in the embedding space.  For the correction task, this means that incorrect recognition hypotheses and phonetically-similar corrections are mapped to nearby points in the embedding space.
    \item At inference, we produce embeddings of the correction target and candidate corrections, and a fast approximate neighbors system (e.g., ScaNN~\cite{scann_paper}) is used to find candidate corrections in the correction candidate database at inference time.  Note that, if the set of candidate corrections is known ahead of time (e.g., as in the case of a media entity database), the embeddings of the candidate corrections can be computed before inference.
\end{enumerate}
ASR correction using nearest neighbor embedding search in large candidate databases was previously demonstrated using the existing top ASR hypothesis's text as the embedder input for retrieving similar phrases from a database of (embedded) textual candidate corrections~\cite{neuralembeddings}.
However, the textual hypothesis may be far from the phonetic ground truth; using a phonetically-inaccurate hypothesis to retrieve nearest neighbors for hypothesis rewriting can lead to low precision, since the search key does not accurately reflect the utterance audio.
To overcome this limitation, we demonstrate a contextual ASR hypothesis correction system based on \emph{multimodal} speech-text embeddings with the following improvements over text-based approaches:

\begin{itemize}
\item \textit{Similar phrase retrieval directly from utterance audio}\\
Our nearest-neighbor retrieval is based on query embeddings computed directly from the utterance audio, eliminating the imprecision that may be introduced by phonetically-inaccurate ASR hypotheses. Further, the encoders that compute these embeddings can be trained specifically to the ASR engine, language, application domain, etc.~\cite{neuralembeddings}

\item \textit{Large-scale, modular, efficient short-form ASR correction}\\
Our approach can be applied on top of a frozen base ASR model. Computing fixed-size embeddings enables efficient matching against a large collection (up to 128K entries in this paper) of contextually-relevant phrases using fast approximate-nearest-neighbors search~\cite{scann_paper}. For voice search, using a single collection simplifies contextual ASR correction, since the phrases can be applied across different contextual situations (e.g., device is playing music, timer active, etc.). More details about how we select such phrases are in Section~\ref{sec:retrieval_candidates}.
\end{itemize}

%% file: related_work.tex
\begin{figure*}[ht]
\centering
\includegraphics[width=12cm]{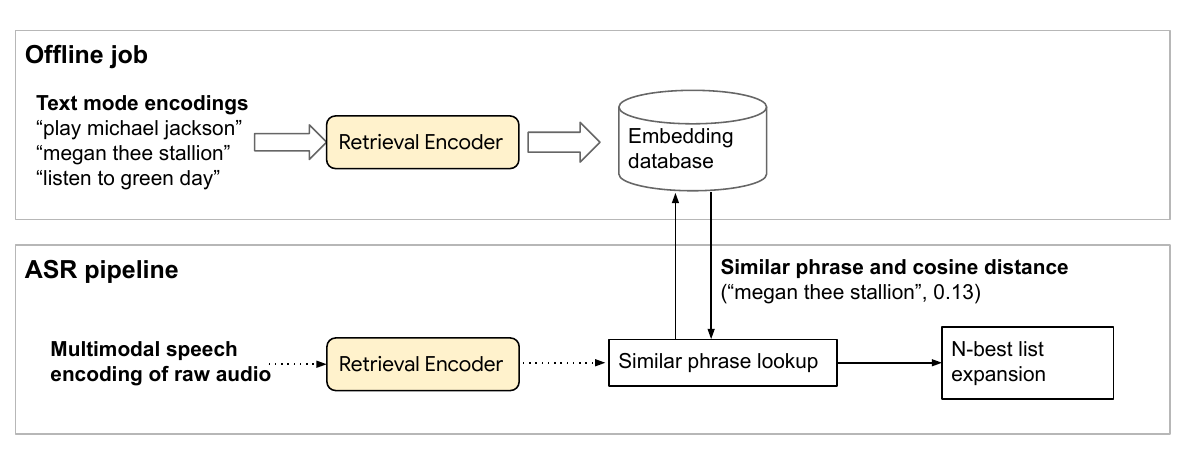}
\caption{Overall system. An offline job builds the embeddings database of retrievable phrases. The utterance audio is used to retrieve nearest neighbors. During N-best list expansion, the nearest neighbor phrase is scored and unioned with the original n-best list.}
% using $n=3$-best hypotheses from ASR outputs.
% In the matrix, diagonal entries measure the cosine similarity between confusable
% entities in the training batch of size $B$=3.}
\label{fig:inference}
\vspace{-0.45cm}
\end{figure*}

\section{Related work}
% General theme of work is increasing recall by reference to external context
% Biasing: classic and E2E
% Spelling correction and rescoring — maybe we can drop the rescoring ref?
% Keyword spotting
% General retrieval
% Specifics of our encoders

% Our approach corrects ASR hypotheses using additional context provided at inference time, in the form of a database of alternative hypotheses.
% %
% Below, we survey, and compare to, other methods of ASR contextualization and biasing.

Classical contextualization approaches for hybrid ASR use weighted finite state transducers
(WFSTs) to bias contextually-relevant phrases by
interpolating their base LM costs with those from a WFST-encoded external language model
(LM) during decoding~\cite{twiddlerpaper}, with optional search space expansion to
increase the likelihood that potentially-biased phrases survive beam pruning~\cite{Williams2017-fx}.
For E2E ASR systems without language models, shallow fusion approaches have been proposed, as in~\cite{shallow_fusion}.
These approaches are limited by the ability to surface contextually-relevant phrases during first-pass beam search, limits on the sizes of WFSTs that can be effectively
constructed and traversed during search, and the lack of language models in pure E2E ASR systems.
% , especially for phrases that are far outside of
% the training data distribution, as well as limits on the sizes of WFSTs that can be effectively
% constructed and traversed during search.
In pure E2E ASR, attention-based biasing approaches~\cite{Pundak2018-fj,Munkhdalai2022-th,Chang2021-zp} avoid the need for a language model for biasing and mitigate search problems by integrating
biasing context directly into the model's acoustic and/or linguistic representations;
however, such approaches also run into scaling limits as the number of phrases grows.
Similarly, integrating k-NN search with language models~\cite{generalizationthroughmemorization} allows adding variable contextually-relevant n-grams at inference time, potentially improving recall at the cost of tighter integration with first-pass ASR.
Our retrieval approach is independent of first-pass beam search, so that phrase recall is
independent of first-pass performance, does not require a tight integration between
biasing and ASR, and scales to hundreds of thousands of phrases using simple mechanisms.

Typical ASR correction models operate on the lattice, using WFST operations in the phonemes space to find correction candidates~\cite{contextualrecovery,classlmwordmapping}, or on ASR hypotheses, e.g., using sequence-to-sequence transducers to correct instances of contextually-relevant phrases~\cite{wang2022contextual}; recent work has extended textual spelling correction with an initial retrieval step~\cite{antonova2023spellmapper}.
However, such approaches are susceptible to mismatches between the (graphemic or phonemic) ASR output and the audio.
Further work extended contextual spelling correction to incorporate acoustic features~\cite{Wang2023-uq}; in contrast to that work, we use fast nearest-neighbors lookup to scale to large correction candidate sets.

Our system builds on the MAESTRO~\cite{maestro} technique for training ASR models in a self-supervised manner using untranscribed speech and unspoken text.
The suitability of mean- and max-pooled MAESTRO embeddings for retrieval was explored in~\cite{wang2023understanding}; we build on this work by training a \textit{retrieval encoder} in the ``dual encoder'' style~\cite{flare_entity_retrieval} which further processes MAESTRO outputs to produce embeddings that increase recall. Note that, while we build on MAESTRO in this work, our technique can be extended to any underlying model that produces joint speech-text representations, such as JOIST~\cite{sainath2022joist} and STPT~\cite{tang2022unified}.

Our retrieval-based architecture
also resembles general keyword spotting~\cite{gales-babel-kws,audhkhasi-e2e-kws,
sainath-convolutional-kws,alvarez-e2e-streaming-kws}.
Like the approach of
Sacchi et al~\cite{sacchi-embedding-open-kws}, we use learned acoustic embeddings and nearest neighbor search to find close matches from an open vocabulary, but our approach differs in that we use a modality-matching model to directly match text correction candidates against the utterance audio;
notably, we are able to do direct audio-to-text matching without using text-to-speech (TTS) synthesis, as is done in~\cite{acoustic-catalog}.

%% Include reference to other retrieval and keyword search work. e.g. IARPA BABEL program

% Various approaches for rescoring exist, such as using external LMs to rescore
% hypotheses on-the-fly during first-pass decoding~\cite{twiddlerpaper,
% shallow_fusion}. In addition, it is possible to incorporate scoring into the
% training process for E2E models such as the recurrent neural network transducer
% (RNN-T). Previous approaches fused an external LM with the RNN-T during training
% ~\cite{cold_fusion, component_fusion, deep_fusion}. More recently, E2E models
% using the hybrid autoregressive transducer (HAT) architecture---whose loss
% function contains separate internal and external LM
% components---motivate the use of external LMs to rescore hypotheses in a
% principled manner~\cite{variani2020hybrid}. In this work, we apply our system
% on top of a base E2E ASR model with the HAT architecture.

% Additionally, it is possible to rescore the n-best hypotheses produced after the
% initial decoding pass. Such techniques include reranking the n-best hypotheses
% by applying binary classification~\cite{maxent} and using information theoretic
% measures~\cite{nbestreordering}.  In addition to rescoring the hypotheses,
% it is also possible to apply attention on the sequence generated from
% first-pass decoding and introduce new hypotheses in a redecoding
% step~\cite{deliberation}. 

%% file: system_design.tex
\section {System Design}
\label{sec:neural_expand}
\subsection{Feature extraction with a pretrained MAESTRO model}
\label{ssec:matchinginembeddings}
\begin{figure}[ht]
\centering
\includegraphics[width=5cm,height=5cm]{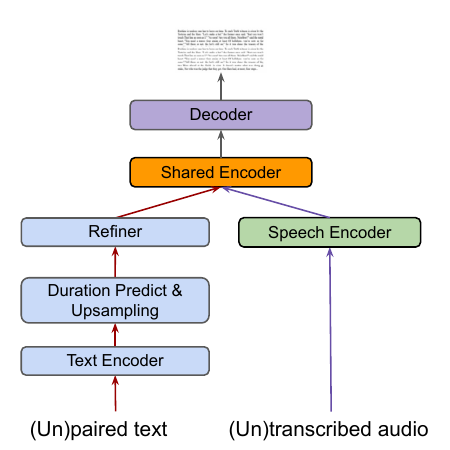}
\caption{MAESTRO model semi-supervised training process.}
\label{fig:text_injection_training}
\vspace{-0.42cm}
\end{figure}

Our system, illustrated in Fig.~\ref{fig:inference}, performs speech-text matching, where the mode of the query (speech) is different from that of the retrievable entity (text). This approach requires a speech-text model that can accurately map spoken text with acoustic variations (in noise level, pitch, duration, etc.) to the ground truth text. We use a MAESTRO model that has has been pretrained on large amounts of both paired and unpaired text-and-speech data to produce speech-text embedding sequences~\cite{maestro}. For our experiments, the underlying MAESTRO model was trained on Librispeech~\cite{librispeech}. Fig.~\ref{fig:text_injection_training} illustrates how the MAESTRO model is trained in a semi-supervised manner.

By design, MAESTRO's neural model is trained on several tasks. The text encoder contains a graphemes-to-phonemes converter, along with a phonemes-upsampler model trained on a duration prediction task using phoneme alignments from an RNN-T decoder. For text inputs, the length of the shared encoder output embedding sequence is based on the duration prediction task, whereas the sequence length for speech inputs is proportional to the audio duration. The speech encoder contains a pretrained w2v-BERT model~\cite{ChungZhangHan21}. A refiner is used to  process the upsampled text inputs during a consistency loss (mean squared error) minimization task between the speech encoder and refiner output for paired data.

MAESTRO contains a shared encoder that accepts as input either speech or text embedding outputs from the corresponding consistency-loss-optimized encoder. The shared encoder---which comprises the final 12 layers out of 24 from a pretrained E2E conformer---outputs to an RNN-T decoder. An additional task involves training the overall architecture on an RNN-T loss function. The total parameter count for the MAESTRO model is 650 million. Of those parameters, 300 million are for the speech model, 300 million are for the shared encoder, and 50 million are for the text model.

\begin{figure}[ht]
\centering
\includegraphics[width=5cm,height=8cm]{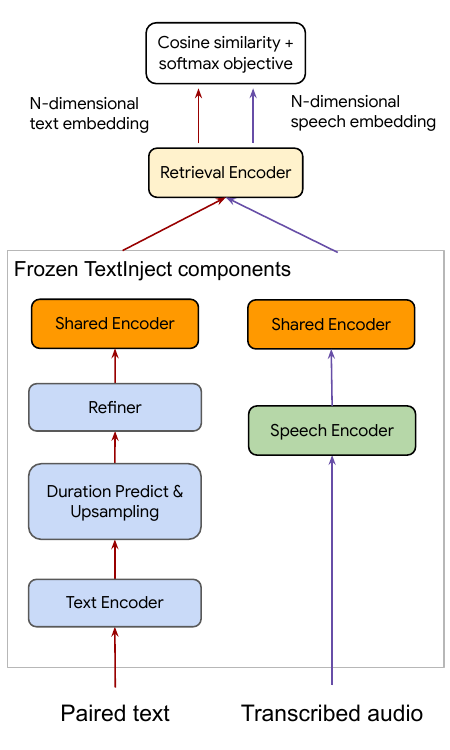}
\caption{Retrieval Encoder supervised training process.}
\label{fig:encoder_training}
\vspace{-0.5cm}
\end{figure}

\subsection{Speech-text embeddings retrieval with a trained encoder}
\label{ssec:encoder_training}
Our system freezes the MAESTRO shared encoder, using it to output speech-text embedding sequences that can be processed for similar phrase retrieval via a \textit{retrieval encoder}. The retrieval encoder mean pools the sequence of embeddings across time and then processes the 1024-dimensional mean-pooled output with a single layer feed forward neural network (FFNN) combined with a ReLU activation with dropout probability 0.1.
\
The resulting 1024-dimensional output embedding represents the phrase. The FFNN contains a total of 2.1 million parameters. During preliminary analysis, omitted for brevity, we observed that this shallow architecture achieved better retrieval performance than more-complex architectures. In Section~\ref{ssec:overall_eval}, we demonstrate better performance using a retrieval encoder, rather than simple mean-pool-only representations, for retrieval.

The retrieval encoder was trained using the dual encoder framework as follows (illustrated in Fig.~\ref{fig:encoder_training}). Consider a batch of $B$ (text, speech) pairs, $(t_i, s_i)$, $i \in [1, B]$.
Let $z(x)=r(\bar{y}(x))$ represent the retrieval encoder output for text or speech input $x \in \{t_i,..., t_B, s_i,...,s_B\}$, where $\bar{y}(x)$ is the mean-pooled shared encoder output on text or speech input $x$.

The encoder is trained with softmax cross entropy loss: $$-\frac{1}{B}\sum_{i=1}^{B} \log{S_{i}}\textrm{,}$$
where $S_{i} = e^{p_{i,i}} /\sum_{j=1}^{B} e^{p_{i,j}}$ is the softmax function applied to a row in a matrix of \textit{speech-text} cosine similarities, $p$, such that $$p_{i,j} = \text{sim}(t_i, s_j) = \frac{z(t_i)\cdot z(s_j)}{\|z(t_i)\| \|z(s_j)\|} \textrm{.}$$

The training data for the retrieval encoder consists of multimodal text-and-speech embedding pairs originating from the LibriTTS data sets~\cite{libritts}, along with an internal collection of 254K manually-transcribed, anonymous short-form human audio utterances of random places, businesses, and voice assistant commands. Together, these pairs formed a training set of 454K examples.

The retrieval encoder was trained on batches with 128 examples over 1.5 million training steps
using mini-batch gradient descent with momentum
(momentum parameter of 0.9)~\cite{gradient_descent}
with a learning rate of 0.001 without decay and 10K warm up steps with an
initial learning rate of 0.0 and a linear increase to the final learning rate. 

%% file: inference.tex
\subsection{Inference with the shared encoder and retrieval encoder}
\label{ssec:acoustic_scoring}
In our system, the shared encoder and retrieval encoder are used to process the query audio $s$ and a collection of $M$ contextually-relevant phrases $t \in {t_1,...,t_M}$, where the embeddings $\{z(t_l)\}_{l=1}^{M}$ are computed offline. Let $\hat{t} = \argmax_{t} \text{sim}(t, s)$ represent the nearest neighbor text phrase returned during inference.

Our system adds $\hat{t}$ to the n-best list with cost $c(\hat{t})$ as a function of the E2E cost of the top hypothesis in the original n-best list $c_\text{orig}$,  a \textit{rewriting aggressiveness} hyperparameter $\delta \in [0, 1]$, and the speech-text similarity as follows:
\begin{equation}
c(\hat{t}) = c_\text{orig} + \left(1 - \text{sim}\left(\hat{t}, s\right)\right) - \delta
\end{equation}

The cost $c(\hat{t})$ can be interpreted as a pseudo negative log-likelihood. Ultimately, if $c(\hat{t}) < c_\text{orig}	\Leftrightarrow 1 - \text{sim}\left(\hat{t}, s\right) < \delta$, then $\hat{t}$ becomes the new top ASR hypothesis. We limited $\delta$ to the range $[0,1]$ since $\delta \geq 1$ would allow new top hypotheses $\hat{t}$ dissimilar to $s$, i.e.,  $\text{sim}(\hat{t},s) < 0$.

While it is possible to union $\hat{t}$ (containing score $c(\hat{t})$) with the original n-best list and then rescore/rerank the expanded hypothesis list with an external LM, we did not observe any meaningful improvement in separate experiments.

%% file: experiments.tex
\section{Experiments}
\label{sec:experiments}
\subsection{Base ASR architecture}
\label{ssec:baseasr}
% Details about embedding decoder: go/embedding-decoder
% Note from rondon@: non-causal encoder not used in server models as of 3/21/22
% During a second pass, the decoder receives
% outputs from the same prediction network along with outputs from a
% non-causal layer processing the left-context and right-context first-pass
% encoder outputs. The non-causal encoder processes input 900 milliseconds into
% the future.
Our E2E ASR system is based on the architecture described in~\cite{e2e}.
The input contains 128-dimensional log-Mel features with a 16-dimensional
one-hot domain-id vector appended. The network uses an embedding decoder
in which the previous 2 output tokens are used to look up an embedding
before applying a projection. A joint layer computes the distribution over output tokens
based on the projected output from the embedding decoder network and the encoder
output. The encoder consists of 12 Conformer layers, each
of which contains an 8-head self-attention layer and a convolutional kernel
with size 15. The overall network (120 million total parameters) predicts tokens from a 4,096-wordpiece
vocabulary~\cite{Schuster2012}. The model was trained using the HAT
factorization~\cite{variani2020hybrid}; inference incorporates a first-pass LM (5-gram model with 4 million word vocabulary) implemented by a WFST whose input tape contains wordpieces and output tape contains written-domain
words~\cite{hybridseq2seq}. 400K hours of anonymized data were used as the training data. When dealing with user data, our work abides by Google AI principles~\cite{aiprinciples}.

\subsection{Description of models}
We compared the following ASR setups in our experiments:

\textbf{Base} corresponds to the base E2E ASR described in Section~\ref{ssec:baseasr}. Each of the setups below corrects outputs from the base ASR.
    
\textbf{Mean Pool} omits the retrieval encoder and computes $z(x) = \bar{y}(x)\textrm{, } \forall x \in \{s, t\}$ during inference and offline computation of $\{z(t_l)\}_{l=1}^{M}$ using only the MAESTRO encoders described in Section~\ref{ssec:matchinginembeddings}. This setup is identical to that used in previous MAESTRO retrieval experiments~\cite{wang2023understanding}.
    
\textbf{Mean Pool+Ret (our system)} is based on the system described in Sections~\ref{ssec:encoder_training} and ~\ref{ssec:acoustic_scoring}.

\textbf{Text+Ret} uses a dual encoder that is trained on (transcript truth, top ASR hypothesis) pairs. As both inputs are text, this setup uses a single shared encoder with a two-layer Transformer architecture. Initially, each phrase is converted to a sequence of character token embeddings (maximum length 100, each 100-dimensional). In each Transformer layer, the input sequence is processed by 4 self-attention heads, which together output a sequence of 400-dimensional embeddings. Each embedding is identically processed by a single layer FFNN that outputs a sequence of 400-dimensional embeddings, forming the Transformer layer output.

The final embedding is equal to the mean pool applied to the final Transformer layer output followed by a projection to 1024 dimensions. This architecture (30.5 million total parameters) is similar to that used in previous text-only neural matching experiments~\cite{neuralembeddings}.

% Implementation details about learning rate:
% http://google3/nlp/neon/dual_encoder/utils/estimator.py?l=736&rcl=437049935

% of 0.01. The size of the trained encoder, including the TensorFlow~\cite{AbadiAgarwalBarhamEtAl15} graph and its associated metadata, is 13 MB.

\subsection{Building the embedding database}
\label{sec:retrieval_candidates}We sampled $M$=128K transcripts from a base collection of labeled utterances contextually relevant for voice search, including acquired human audio voice search queries for directions to locations and TTS queries for application control, dictation, calling contacts, and media synthesized from grammars. The transcript text was then encoded using the retrieval encoder during addition to the embedding database. The offline job in Figure~\ref{fig:inference} illustrates this procedure.

\subsection{Test sets}
Throughout our experiments, we evaluate on two types of test sets, each with dev and eval splits. The first is an \textit{in-context} (IC) test set, where each transcript truth is present in the embedding database as a retrievable phrase. The purpose of this test set is to measure the recall of an ASR correction system. The \textbf{IC} set contained 4K utterances from the base collection described in Section~\ref{sec:retrieval_candidates}.

The second is an \textit{anti} test set, which is intended to measure precision. This test set is designed to represent background voice search queries, i.e., those to which no retrievable correction applies. Transcript truths from these utterances are not added to the embedding database. For a contextual ASR system, performance on such queries ideally should be no worse than the Base model. 
% Table~\ref{table:testsets} summarizes the test sets in our experiments.
\textbf{AntiTTS} contained 5K TTS-synthesized popular generic voice assistant queries. In addition, \textbf{AntiH} (used during eval split only) contained 9K anonymized human audio voice search queries.

All test sets contain short-form utterances of a few words reflective of common voice search use cases, e.g., ``places to eat near me,'' ``what's the weather on Friday?''

%% file: results.tex
\newcommand{\tworow}[1]{\multirow{2}{*}{#1}}
\newcommand{\meanpool}{\tworow{MeanPool}}
\newcommand{\maxpool}{\tworow{MaxPool}}
\newcommand{\meanpoolret}{\tworow{MeanPool+Ret}}
\newcommand{\maxpoolret}{\tworow{MaxPool+Ret}}
\newcommand{\textret}{\tworow{Tex+Ret}}
\newcommand{\deltamstar}{$\delta_{m}^{*}$}

\section{Results}
\label{ssec:overall_eval}
We evaluated on the eval split utterances while increasing $M$ to measure the quality of embeddings when searching against increasingly-large embedding databases. For each model $v\in\{\text{Mean Pool}, \text{Mean Pool+Ret}, \text{Text+Ret}\}$, we swept the model-specific rewriting aggressiveness $\delta_v^{*} \in [0, 1]$ to find the value producing the lowest weighted WER score on the IC and AntiTTS dev splits for $M$=8K, with selected weights: 95\% anti, 5\% in-context. The dev-split-determined hyperparameter values for different were then used when evaluating on the eval split. 

The results for varying $M$ are shown in Table~\ref{table:results_vs_embeddings_database_size}.  For the $M\in\{$8K, 16K, 32K, 64K$\}$ settings, MeanPool+Ret achieved the lowest in-context WER without increasing anti set WER relative to Base. On the other hand, Text+Ret's AntiTTS WER increased with $M$, suggesting reduced precision likely due to querying using embeddings of ASR hypotheses that were phonetically dissimilar to the ground truth, but confusable with the ASR correction candidates in the embedding database. For all models, WER on AntiH did not increase relative to the Base AntiH WER. This suggests that our embedding databases were more likely to contain phrases from domains different from those in AntiH's utterances.

\begin{table}
\caption{Eval WER values for increasing embedding database sizes.}
\centering
\label{table:results_vs_embeddings_database_size}
\scalebox{0.9}{
\begin{tabular}{lcrrrrr}\toprule
\tworow{Model} & \tworow{Test set} &  \multicolumn{5}{c}{Embedding database size $M$} \\
& & 8K & 16K & 32K & 64K & 128K \\
\midrule
\tworow{Base} & IC & 14.9 & 14.9 & 14.9 & 14.9 & 14.9 \\
& AntiTTS & \textbf{2.4} & \textbf{2.4}  & \textbf{2.4} & \textbf{2.4}  & \textbf{2.4} \\
& AntiH & \textbf{6.2} & \textbf{6.2} & \textbf{6.2} & \textbf{6.2} & \textbf{6.2} \\

\midrule
\tworow{Text+Ret} & IC & 13.9 & 13.9 & 13.9 & 13.9  & \textbf{13.9} \\
& AntiTTS & \textbf{2.4} & 2.5 & 2.6 & 2.7 & 2.9 \\
& AntiH & \textbf{6.2} & \textbf{6.2} & \textbf{6.2} & \textbf{6.2} & \textbf{6.2} \\
 \midrule
\meanpool & IC & 14.7 & 14.7 & 14.7 & 14.7 & 14.7\\
& AntiTTS & \textbf{2.4} & \textbf{2.4} & \textbf{2.4} & \textbf{2.4} & \textbf{2.4} \\
& AntiH & \textbf{6.2} & \textbf{6.2} & \textbf{6.2} & \textbf{6.2} & \textbf{6.2}  \\
\midrule
\meanpoolret & IC & \textbf{13.6} & \textbf{13.7} & \textbf{13.8} & \textbf{13.8} & 14.0 \\
& AntiTTS & \textbf{2.4} & \textbf{2.4} & \textbf{2.4} & \textbf{2.4} & \textbf{2.4}\\
& AntiH &  \textbf{6.2} & \textbf{6.2} & \textbf{6.2} & \textbf{6.2} & \textbf{6.2} \\

\bottomrule
\end{tabular}
} % scalebox
\end{table}

%% file: conclusion.tex
\section{Conclusion}
\label{sec:conclusion}
We introduced a system where similar phrases are retrieved via a nearest neighbor search with speech-text embeddings to match the utterance audio with candidate text phrases. This enabled ASR correction with greater precision than text-based retrieval, which may introduce implausible phrases due to inputs that are too phonetically dissimilar from the ground truth. We achieved a relative word error rate reduction of 6\% on a voice search test set containing transcript truths that are included in a database of 128K retrievable phrases without degrading recognition of general utterances.

%% file: acknowledgements.tex
\section{Acknowledgements}
\label{sec:acknowledgements}
The authors would like to thank Pavel Golik, Richard Rose, and Parisa Haghgani for their insightful
comments and discussions.

%% file: main.bbl
\begin{thebibliography}{10}

\bibitem{flare_entity_retrieval}
D.~Gillick, S.~Kulkarni, L.~Lansing, A.~Presta, J.~Baldridge, E.~Ie, and
  D.~Garcia-Olano,
\newblock ``Learning dense representations for entity retrieval,''
\newblock in {\em Conference on Computational Natural Language Learning
  (CoNLL)}, 2019.

\bibitem{scann_paper}
R.~Guo, P.~Sun, E.~Lindgren, Q.~Geng, D.~Simcha, F.~Chern, and Sanjiv Kumar,
\newblock ``Accelerating large-scale inference with anisotropic vector
  quantization,''
\newblock in {\em International Conference on Machine Learning (ICML)}, 2020.

\bibitem{neuralembeddings}
C.~Li, P.~Rondon, D.~Caseiro, L.~Velikovich, X.~Velez, and P.~Aleksic,
\newblock ``Improving entity recall in automatic speech recognition with neural
  embeddings,''
\newblock in {\em IEEE International Conference on Acoustics, Speech and Signal
  Processing (ICASSP)}, 2021.

\bibitem{twiddlerpaper}
P.~Aleksic, M.~Ghodsi, A.~Michaely, C.~Allauzen, K.~Hall, B.~Roark, D.~Rybach,
  and Pedro Moreno,
\newblock ``Bringing contextual information to {G}oogle speech recognition,''
\newblock in {\em Interspeech}, 2015.

\bibitem{Williams2017-fx}
I.~Williams and P.~Aleksic,
\newblock ``Rescoring-aware beam search for reduced search errors in contextual
  automatic speech recognition,''
\newblock in {\em Interspeech}, 2017.

\bibitem{shallow_fusion}
D.~Zhao, T.~Sainath, D.~Rybach, P.~Rondon, D.~Bhatia, B.~Li, and Ruoming Pang,
\newblock ``Shallow-fusion end-to-end contextual biasing,''
\newblock in {\em Interspeech}, 2019.

\bibitem{Pundak2018-fj}
G.~Pundak, T.~Sainath, R.~Prabhavalkar, A.~Kannan, and D.~Zhao,
\newblock ``Deep context: End-to-end contextual speech recognition,''
\newblock in {\em {SLT}}, 2018.

\bibitem{Munkhdalai2022-th}
T.~Munkhdalai, K.C. Sim, Angad Chandorkar, F.~Gao, M.~Chua, T.~Strohman, and
  Fran{\c c}oise Beaufays,
\newblock ``Fast contextual adaptation with neural associative memory for
  {On-Device} personalized speech recognition,''
\newblock in {\em ICASSP}, 2022.

\bibitem{Chang2021-zp}
F.~Chang, J.~Liu, M.~Radfar, A.~Mouchtaris, M.~Omologo, A.~Rastrow, and
  S.~Kunzmann,
\newblock ``Context-aware transformer transducer for speech recognition,''
\newblock in {\em {IEEE} Automatic Speech Recognition and Understanding
  Workshop ({ASRU})}, 2021.

\bibitem{generalizationthroughmemorization}
U.~Khandelwal, O.~Levy, D.~Jurafsky, L.~Zettlemoyer, and M.~Lewis,
\newblock ``Generalization through memorization: Nearest neighbor language
  models,''
\newblock in {\em ICLR}, 2020.

\bibitem{contextualrecovery}
J.~Serrino, L.~Velikovich, P.~Aleksic, and C.~Allauzen,
\newblock ``Contextual recovery of out-of-lattice named entities in automatic
  speech recognition,''
\newblock in {\em Interspeech}, 2019.

\bibitem{classlmwordmapping}
R.~Huang, O.~Abdel-hamid, X.~Li, and G.~Evermann,
\newblock ``Class {LM} and word mapping for contextual biasing in end-to-end
  {ASR},''
\newblock in {\em Interspeech}, 2020.

\bibitem{wang2022contextual}
X.~Wang, Y.~Liu, J.~Li, V.~Miljanic, S.~Zhao, and H.~Khalil,
\newblock ``Towards contextual spelling correction for customization of
  end-to-end speech recognition systems,'' 2022.

\bibitem{antonova2023spellmapper}
A.~Antonova, E.~Bakhturina, and B.~Ginsburg,
\newblock ``{SpellMapper}: A non-autoregressive neural spellchecker for {ASR}
  customization with candidate retrieval based on n-gram mappings,''
\newblock in {\em Interspeech}, 2023.

\bibitem{Wang2023-uq}
X.~Wang, Y.~Liu, J.~Li, and S.~Zhao,
\newblock ``Improving contextual spelling correction by external acoustics
  attention and semantic aware data augmentation,''
\newblock in {\em ICASSP}, 2023.

\bibitem{maestro}
Z.~Chen, Y.~Zhang, A.~Rosenberg, B.~Ramabhadran, P.~Moreno, A.~Bapna, and
  H.~Zen,
\newblock ``{MAESTRO}: Matched speech text representations through modality
  matching,''
\newblock in {\em Interspeech}, 2022.

\bibitem{wang2023understanding}
G.~Wang, K.~Kastner, A.~Bapna, Z.~Chen, A.~Rosenberg, B.~Ramabhadran, and
  Y.~Zhang,
\newblock ``Understanding shared speech-text representations,''
\newblock in {\em ICASSP}, 2023.

\bibitem{sainath2022joist}
T.~Sainath, R.~Prabhavalkar, A.~Bapna, Y.~Zhang, Z.~Huo, Z.~Chen, B.~Li,
  W.~Wang, and T.~Strohman,
\newblock ``{JOIST}: A joint speech and text streaming model for {ASR},''
\newblock in {\em SLT}, 2022.

\bibitem{tang2022unified}
Y.~Tang, H.~Gong, N.~Dong, C.~Wang, W.~Hsu, J.~Gu, A.~Baevski, X.~Li,
  A.~Mohamed, M.~Auli, and J.~Pino,
\newblock ``Unified speech-text pre-training for speech translation and
  recognition,''
\newblock in {\em Proceedings of the 60th Annual Meeting of the Association for
  Computational Linguistics (Volume 1: Long Papers)}, 2022.

\bibitem{gales-babel-kws}
M.~J.~F. Gales, K.~M. Knill, A.~Ragni, and S.~P. Rath,
\newblock ``Speech recognition and keyword spotting for low resource langauges:
  Babel project research at {CUED},''
\newblock in {\em International Workshop on Spoken Language Technologies for
  Under-Resourced Languages}, 2014.

\bibitem{audhkhasi-e2e-kws}
K.~{Audhkhasi}, A.~{Rosenberg}, A.~{Sethy}, B.~{Ramabhadran}, and
  B.~{Kingsbury},
\newblock ``End-to-end {ASR}-free keyword search from speech,''
\newblock {\em IEEE Journal of Selected Topics in Signal Processing}, vol. 11,
  no. 8, pp. 1351--1359, 2017.

\bibitem{sainath-convolutional-kws}
T.~Sainath and C.~Parada,
\newblock ``Convolutional neural networks for small-footprint keyword
  spotting,''
\newblock in {\em Interspeech}, 2015.

\bibitem{alvarez-e2e-streaming-kws}
R.~{Alvarez} and {H}.-{J}. {Park},
\newblock ``End-to-end streaming keyword spotting,''
\newblock in {\em ICASSP}, 2019.

\bibitem{sacchi-embedding-open-kws}
N.~Sacchi, A.~Nanchen, M.~Jaggi, and M.~Cernak,
\newblock ``{Open-Vocabulary Keyword Spotting with Audio and Text
  Embeddings},''
\newblock in {\em Interspeech}.

\bibitem{acoustic-catalog}
D.M. Chan, S.~Ghosh, A.~Rastrow, and B.~Hoffmeister,
\newblock ``Domain adaptation with external off-policy acoustic catalogs for
  scalable contextual end-to-end automated speech recognition,''
\newblock in {\em ICASSP}, 2023.

\bibitem{librispeech}
V.~Panayotov, G.~Chen, D.~Povey, and S.~Khudanpur,
\newblock ``Librispeech: An {ASR} corpus based on public domain audio books,''
\newblock in {\em ICASSP}, 2015.

\bibitem{ChungZhangHan21}
Y.-A. Chung, Y.~Zhang, W.~Han, C.-C. Chiu, J.~Qin, R.~Pang, and Yonghui Wu,
\newblock ``w2v-bert: Combining contrastive learning and masked language
  modeling for self-supervised speech pre-training,''
\newblock in {\em ASRU}, 2021.

\bibitem{libritts}
H.~Zen, V.~Dang, R.~Clark, Y.~Zhang, R.~J. Weiss, Y.~Jia, Z.~Chen, and Y.~Wu,
\newblock ``{LibriTTS}: A corpus derived from {LibriSpeech} for
  text-to-speech,''
\newblock in {\em Interspeech}, 2019.

\bibitem{gradient_descent}
I.~Sutskever, J.~Martens, G.~Dahl, and G.~Hinton,
\newblock ``On the importance of initialization and momentum in deep
  learning,''
\newblock in {\em ICML}, 2013.

\bibitem{e2e}
T.~Sainath, Y.~He, A.~Narayanan, R.~Botros, R.~Pang, D.~Rybach, C.~Allauzen,
  E.~Variani, J.~Qin, Q.-N. Le-{T}he, A.~Gruenstein, A.~Gulati, B.~Li,
  C.~Peyser, C.-C. Chiu, D.~Caseiro, E.~Guzman, I.~McGraw, J.~Yu, M.D. Riley,
  P.~Rondon, Q.~Liang, S.~Mavandadi, S.-{Y}. Chang, T.~Strohman, W.~R. Huang,
  W.~Li, Y.~Wu, and Y.~Zhang,
\newblock ``An efficient streaming non-recurrent on-device end-to-end model
  with improvements to rare-word modeling,''
\newblock in {\em Interspeech}, 2021.

\bibitem{Schuster2012}
M.~Schuster and K.~Nakajima,
\newblock ``Japanese and {K}orean voice search,''
\newblock {\em ICASSP}, 2012.

\bibitem{variani2020hybrid}
E.~Variani, D.~Rybach, C.~Allauzen, and M.~Riley,
\newblock ``Hybrid autoregressive transducer (hat),''
\newblock in {\em ICASSP}, 2020.

\bibitem{hybridseq2seq}
C.~Allauzen, E.~Variani, M.~Riley, D.~Rybach, and H.~Zhang,
\newblock ``A hybrid seq-2-seq {ASR} design for on-device and server
  applications,''
\newblock in {\em Interspeech}, 2021.

\bibitem{aiprinciples}
Google,
\newblock ``Artificial intelligence at {G}oogle: Our principles,''
  \url{https://ai.google/principles}.

\end{thebibliography}
